\documentclass[runningheads]{llncs}
\usepackage{graphicx}
\usepackage{subcaption}
\usepackage{xcolor, cite}
\usepackage{amsmath,amsfonts,amssymb,dsfont}
\usepackage[misc]{ifsym}
 
\usepackage{amsfonts}
\usepackage{nicefrac}

\usepackage{amsmath}
\usepackage{graphicx}
\usepackage{cite}

\usepackage{microtype}
\usepackage{graphicx}
\usepackage{booktabs}
\usepackage{multirow}
\usepackage{xcolor}

\usepackage{isomath}

\usepackage{bm}
\usepackage{mathtools}
\newcommand{\mfor}[1]{\bm{\mathit{#1}}}
\newcommand{\vfor}[1]{\mathbf{#1}}

\def\R{\mathbb{R}}

\def\our{SGCN}
\def\relu{\mathrm{ReLU}}
\def\mlp{\mathrm{MLP}}

\def\R{\mathbb{R}}

\def\relu{\mathrm{ReLU}}
\def\mlp{\mathrm{MLP}}
\def\1{\mathds{1}}

\begin{document}
\title{Processing of incomplete images by (graph) convolutional neural networks\thanks{A preliminary version of this paper appeared as an extended abstract \cite{danel2020processing} at the ICML Workshop on The Art of Learning with Missing Values.}}
\titlerunning{Learning from Incomplete Images Using (Graph) CNNs}
\author{Tomasz Danel\inst{1} \and Marek \'Smieja\inst{1} \and \L{}ukasz  Struski\inst{1} \and
Przemys\l{}aw Spurek\inst{1} \and \L{}ukasz Maziarka\inst{1}}
\authorrunning{T. Danel et al.}
\institute{Faculty of Mathematics and Computer Science\\
Jagiellonian University\\
\L{}ojasiewicza 6, 30-428 Krakow, Poland\\
\email{\{tomasz.danel, marek.smieja\}@ii.uj.edu.pl}}
\maketitle              
\begin{abstract}
We investigate the problem of training neural networks from incomplete images without replacing missing values. For this purpose, we first represent an image as a graph, in which missing pixels are entirely ignored. The graph image representation is processed using a spatial graph convolutional network (\our{}) -- a type of graph convolutional networks, which is a proper generalization of classical CNNs operating on images. On one hand, our approach avoids the problem of missing data imputation while, on the other hand, there is a natural correspondence between CNNs and \our{}.
Experiments confirm that our approach performs better than analogical CNNs with the imputation of missing values on typical classification and reconstruction tasks.

\end{abstract}

\section{Introduction}

Learning from missing data is one of the basic challenges in machine learning and data analysis \cite{goodfellow2016deep}. In a typical pipeline, missing data are first replaced by some values (imputation) and next the complete data are used for training a given machine learning model \cite{mcknight2007missing}. The above approach depends strictly on the imputation procedure -- if we accurately predict missing values, then the other model that operates on completed inputs can obtain good performance.
However, it is not obvious how to select imputation method for a given problem, because it is difficult to validate its performance in a real-life scenario. Thus, there appears a natural question: \emph{can we learn from missing data directly without using any imputation at the preprocessing stage?}

While it is difficult to answer this problem in general, a few approaches have already been designed for particular machine learning models \cite{dekel2010learning, globerson2006nightmare}. In \cite{chechik2008max} a modified SVM classifier is trained by scaling the margin according to observed features only. In \cite{grangier2010feature}, the embedding mapping of feature-value pairs is constructed together with a classification objective function. Pelckmans et al. \cite{pelckmans2005handling} model the expected risk, which takes into account the uncertainty of the predicted outputs when missing values are involved. In a similar spirit, a random forest classifier is modified to adjust the voting weights of each tree by estimating the influence of missing data on the decision of the tree \cite{xia2017adjusted}. The authors of \cite{hazan2015classification} design an algorithm for kernel classification that performs comparably to the classifier which has access to complete data. Goldberg et al. \cite{goldberg2010transduction} treat class labels as an additional column in the data matrix and fill missing entries by matrix completion. The work \cite{smieja2018processing} shows how to generalize fully connected neural networks to the case of missing data given only an imprecise Gaussian estimate of missing data. In the similar spirit, RBF kernel can be calculated for missing data \cite{smieja2019generalized}. Liu et al. \cite{liu2018image} introduce partial convolutions, where the convolution is masked and renormalized to be conditioned on only observed pixels.

In this paper we interpret the image as a graph, in which each node coincides with a visible pixel, while edges connect neighboring pixels, see Figure \ref{fig:graph}. Since missing values are not mapped to graph nodes, we avoid the problem of missing data imputation. In order to efficiently process such an image representation, we use spatial graph convolutional neural networks (\our{}) \cite{spurek2019geometric}. In contrast to typical graph convolutions \cite{kipf2016semi, velivckovic2017graph}, which consider graph as a relational structure invariant to rotations and translations, \our{} introduces a theoretically-justified mechanism to take into account spatial coordinates of nodes. 
More precisely, it has been proven that any layer of convolutional neural networks (CNNs) can be represented as a spatial graph convolution. This fact allows us to think about \our{} as a generalization of CNNs, which is able, in particular, to process incomplete images without imputation.

To verify the introduced procedure, we consider MNIST \cite{lecun1998gradient} and SVHN \cite{netzer2011reading} image datasets. Experimental results show that \our{} performs better than typical CNNs with imputations on the tasks of image classification and reconstruction. 

\section{Graph-based model for processing incomplete images}

In this section, we introduce our model for processing incomplete images. First, we describe how to interpret images as graphs. Next, we recall basic idea of graph convolutional networks (GCNs). Finally, we show the construction of \our{} and discuss it from an intuitive point of view.

\subsection{General idea} 

Images can be interpreted as vectors (tensors) of fixed sizes. If the values of selected pixels are unknown, then the vector structure is destroyed. To recover this structure, we need to replace missing attributes with some values. Substituting unknown inputs carries the risk of introducing unreliable information and noise to initial data. This may have negative consequences on data interpretation as well as can decrease the performance of subsequent machine learning algorithms applied to completed inputs.

Our idea is to interpret an incomplete image as a graph. Graphs represent a relational structure, in which the number of nodes and edges is not fixed. If some pixels in the image are unknown, then the corresponding graph contains less nodes, but the way it is processed does not change. In consequence, graph-based representation of incomplete images is more natural than using imputation.

It is well-known that CNNs are state-of-the-art feature extractors for images. However, as explained above, it is not obvious how to apply CNNs to incomplete data without replacing missing values. In this paper, we use \our{}, which is a type of graph convolutional networks, that takes spatial coordinates of nodes into account. It has been proven that \our{} can mimic any image convolution and, in consequence, \our{} is able to work comparably to CNNs using analogical network architecture (number of layers, size of filters, etc.).

\subsection{Graph-based representation of incomplete images} 

Formally, the image is represented as a tensor $\tensorsym{H} = (\vfor{h}_{ijk}) \in \R^{n \times m \times l}$, where $n,m$ denote height and width of the image, and $l$ is the number of channels. In the case of missing data, we do not have information about pixels values at some coordinates. Thus the incomplete image is denoted by a pair $(\tensorsym{H}, J)$, where $J \subset \{1,\ldots,n\} \times \{1,\ldots,m\}$ indicates pixels which are unknown. In other words, $\vfor{h}_{ijk}$ is unknown for every $(i,j) \in J$. Clearly, for a fully-observed image, $J = \emptyset$.

\begin{figure}[t]
    \centering
    \includegraphics[width=0.18\textwidth]{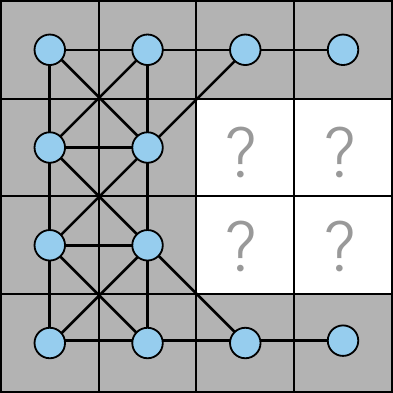}
    \caption{Graph construction for an incomplete image of the size $4 \times 4$ with a missing region of the size $2 \times 2$.}
    \label{fig:graph}
\end{figure}

To construct a graph-based image representation, we create a node for every visible pixel of $\tensorsym{H}$, i.e. 
$$
V = \{v_{ij}: (i,j) \in J'\},
$$
where $J'$ is the set of indices of the observed components. The edge is defined only for nodes that represent adjacent pixels. Formally,
$$
E = \{(v_{ij}, v_{pq}):(i,j)-(p,q) \in \{-1,0,1\}^2 \}.
$$
Observe that for a complete image, every ``non-boundary'' pixel (node) has exactly 8 neighbors. In the case of incomplete data, the number of neighbors can be smaller, as the unknown pixels are not converted to nodes and, in consequence, the corresponding edge is not created, see Figure \ref{fig:graph}. The information about pixels brightness is supplied with a feature vector $\vfor{h}_{ij} \in \R^l$ that corresponds to a node:
$$
\tensorsym{H} = \{\vfor{h}_{ij}: (i,j) \in J'\}.
$$
For a gray-scale image, $\vfor{h}_{ij} \in \R$, while for a color picture $\vfor{h}_{ij} \in \R^3$.  

\subsection{Graph convolutions} 

Let $G=(V,E,\tensorsym{H})$ be a graph (representing the image $\tensorsym{H}$) with $n$ nodes. To avoid multiple indexes in the following description, the node and the corresponding feature vector are denoted by $v_i$ and $\vfor{h}_i$, respectively, while $e_{ij}$ is the edge between $v_i$ and $v_j$. To make a natural correspondence between graphs and images, we put $\vfor{i}=\begin{pmatrix}i_x\\i_y\end{pmatrix}$ to denote both pixel coordinates and index in graph.

\begin{figure}[t]
    \centering
    \includegraphics{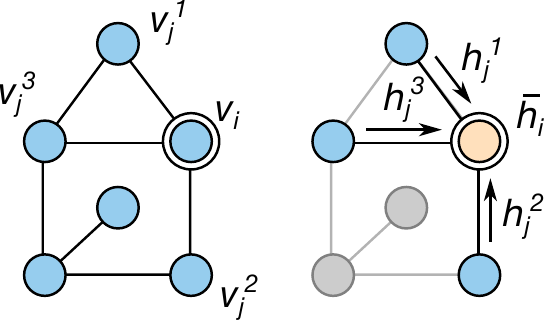}
    \caption{Basic idea of GCNs. Every filter is responsible for defining a pattern used to aggregate feature vectors from adjacent nodes.}
    \label{fig:conv}
\end{figure}

Basic idea of GCNs is to aggregate the information of feature vectors from neighboring nodes over multiple layers, see Figure \ref{fig:conv}. To build a diverse set of patterns, GCNs use filters for defining a specific aggregation. Information from higher-level neighborhoods are fused by combining many layers together.

The above goal is realized by combining two operations. For each node $v_i$, feature vectors of its neighbors are first aggregated:
\begin{equation}\label{eq:neigh}
    \bar{\vfor{h}}_i = \sum_{(v_i,v_j) \in E} u_{ij} \vfor{h}_j.
\end{equation}
Observe that the aggregation is performed only over neighbor nodes, i.e $(v_i,v_j) \in E$. The weights $u_{ij}$ are either trainable \cite{velivckovic2017graph} or determined from a graph \cite{kipf2016semi}. Next, a standard MLP is applied to transform the intermediate representation $\bar{\mfor{H}} = [\bar{\vfor{h}}_1.\ldots,\bar{\vfor{h}}_n]$ into the final output of a given layer:
\begin{equation}\label{eq:mlp}
    \mlp(\bar{\mfor{H}}; \mfor{W}) = \relu(\mfor{W}^T \bar{\mfor{H}} + \vfor{b}),
\end{equation}
where $\matrixsym{W} \in \mathbb{R}^{I\times O}$ is a trainable weight matrix, $\vectorsym{b}\in\mathbb{R}^{O}$ is a trainable bias vector (added column-wise), and $I$ and $O$ is the size of input and output layer respectively.
A typical GCN is composed of a sequence of graph convolutional layers (described above). Finally, its output is aggregated to the network response using a global pooling or a dense layer depending on a given task, e.g. node or graph classification.

\subsection{Spatial graph convolutions} 

\begin{figure}[t]
    \centering
    \includegraphics[width=0.38\textwidth]{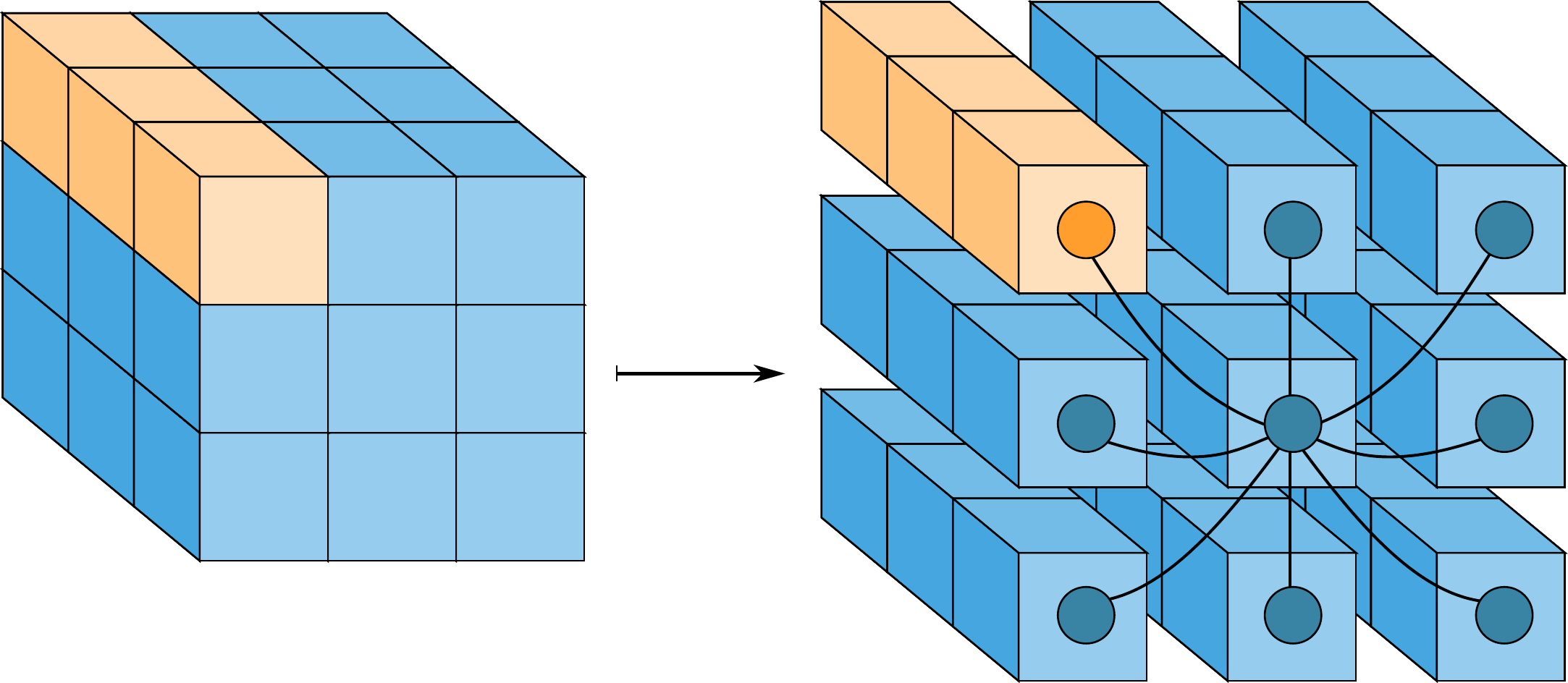}
    \caption{Convolutional kernel for images (left) can be translated to spatial graph convolutions. Vectors at different positions (see e.g. the orange vector in the figure) are multiplied by different weights. In spatial graph convolutions (right) weights are modified by relative positions of graph neighbors to achieve different weights for each vector.}
    \label{fig:geom}
\end{figure}

In contrast to typical GCNs described above, \our{} uses spatial coordinates of nodes, see Figure \ref{fig:geom}. In the case of images, spatial coordinates allows to identify a given pixel in the image grid, which is not possible using only the information about neighborhood. What is more important, the convolution defined by \our{} is constructed so that it is able to reflect any convolutional filter of typical CNNs. In other words, any image convolution can be obtained by a specific parametrization of \our{}. This makes a natural correspondence between \our{} and CNNs. This property cannot be obtained by simply adding spatial coordinates to feature vectors in classical GCNs.

From a formal side, \our{} replaces \eqref{eq:neigh} by:
\begin{equation} \label{eq:neigh-our}
    \bar{\vfor{h}}_i(\mfor{U},\vfor{b}) = \hspace{-4mm}\sum_{(v_i,v_j)\in E}\hspace{-3mm} \relu\left(\mfor{U}\left[\begin{pmatrix}j_x\\j_y\end{pmatrix} - \begin{pmatrix}i_x\\i_y\end{pmatrix}\right] + \vfor{b}\right) \odot \vfor{h}_j,
\end{equation}
where $\mfor{U} \in \R^{I\times2}$, $\vfor{b} \in \R^I$ are trainable, and $I$ is the dimension of the previous layer vectors. The operator $\odot$ is element-wise multiplication. The relative positions in the neighborhood are transformed using a linear operation combined with non-linear ReLU function. This is used to weight the feature vectors $\vfor{h}_j$ in a neighborhood. By the analogy with classical convolution, this transformation can be extended to multiple filters. Let $\tensorsym{U} = [\mfor{U}_1,\ldots,\mfor{U}_k]$ and $\mfor{B} = [\vfor{b}_1,\ldots,\vfor{b}_k]$ define $k$-filters. The intermediate representation $\bar{\vfor{h}}_i$ is a vector defined by:
$$
\bar{\mfor{H}}_i = \left[\bar{\vfor{h}}_i(\mfor{U}_1,\vfor{b}_1), \ldots, \bar{\vfor{h}}_i(\mfor{U}_k,\vfor{b}_k)\right].
$$
Finally, MLP transformation is applied in the same manner as in \eqref{eq:mlp} to transform these feature vectors.

\begin{figure}[h]
\normalsize
\begin{center}
\begin{subfigure}[b]{0.45\textwidth}
\includegraphics[width=0.8\textwidth]{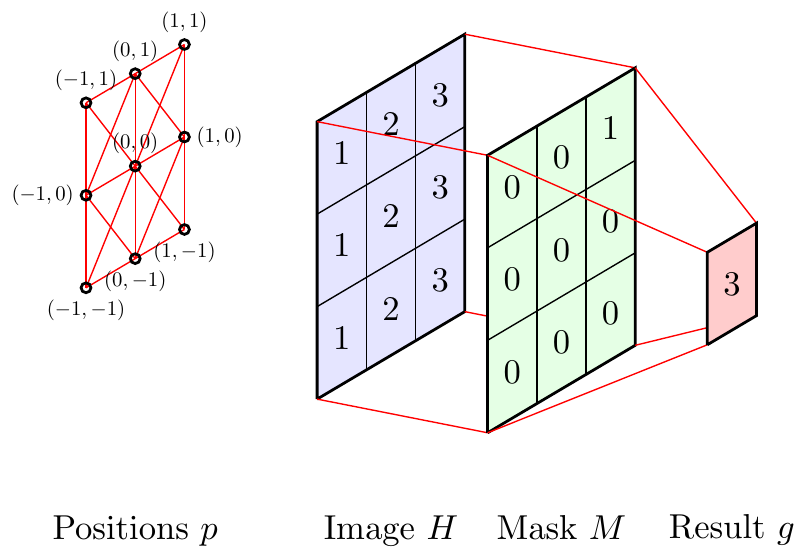} 
\caption{}
\end{subfigure}
\begin{subfigure}[b]{0.35\textwidth}
\includegraphics[width=0.8\textwidth]{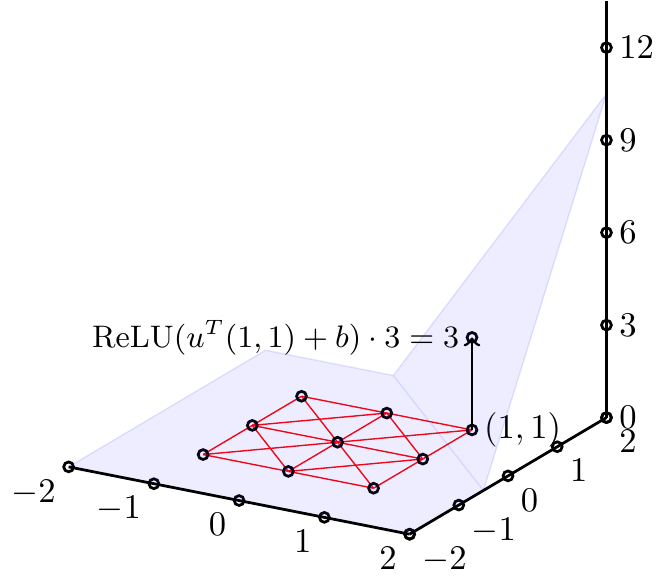} 
\vspace{0.5cm}
\caption{}
\end{subfigure}
\begin{subfigure}[b]{0.44\textwidth}
\includegraphics[width=0.9\textwidth]{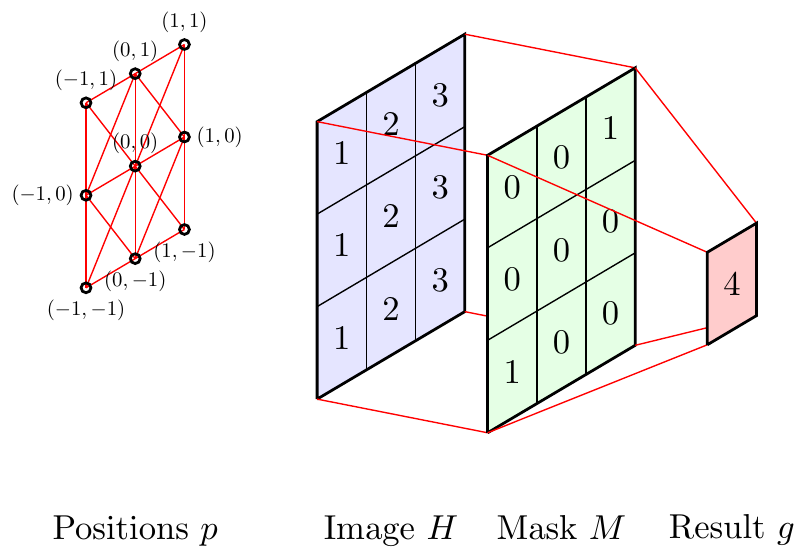}
\caption{}
\end{subfigure}
\begin{subfigure}[b]{0.55\textwidth}
\includegraphics[width=0.48\textwidth]{assets/gr2a-cdot.pdf}
\hspace{0.1cm}
\includegraphics[width=0.48\textwidth]{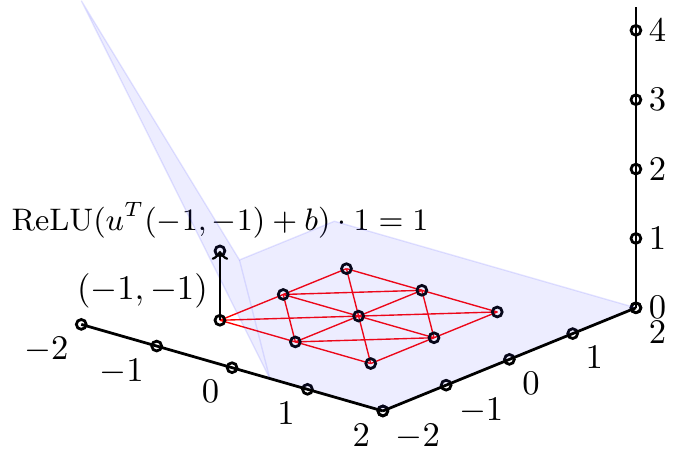}
\vspace{0.4cm}
\caption{}
\end{subfigure}
\end{center}
\caption{Replicating two convolution operations by \our{} (top and bottom), see Examples \ref{ex1} and \ref{ex2} for details. On the left, the result of applying a convolutional filter $\tensorsym{M}$ to the image $\tensorsym{H}$. The positions grid $p$ represents spatial coordinates of the pixels; the neighbors are connected with edges. Analogous convolution can be applied to a spatial graph representation, as shown on the right. In the first case (top) ReLU applied to the linear transformation of the spatial features of the image graph (with $\vfor{u} = (2,2)^T$ and $b=-3$) allows to select (and possibly modify) the top-right neighbor. In the second case (bottom), convolution operation can be obtained by extracting two opposite corner values (with $\vfor{u}_1=(2,2)^T, b_1=-3$ and $\vfor{u}_2=(-2,-2)^T, b_2= -3$) and summing them.
} \label{fig:ex1}
\end{figure}

\subsection{Intuition behind Spatial Graph Convolutions} 

As mentioned, the formulas of \our{} allow to imitate the filters of classical CNNs. While the formal proof of this fact can be found in \cite{spurek2019geometric}, in this section, we demonstrate this property on toy examples, which help to better understand the reasoning behind \our{}. Let us recall that the classical convolution operation (without pooling) defined by the mask $\tensorsym{M}=(m_{i'j'})_{i',j' \in \{-k..k\}}$ applied to the image $\tensorsym{H}=(h_{ij})_{i \in \{1..n\}, j \in \{1..m\}}$ is given by
$$
\tensorsym{M}*\tensorsym{H}=\tensorsym{G}=(g_{ij})_{i \in \{1..n\},j \in \{1..m\}},
$$
where
$$
g_{ij}=\sum_{\begin{array}{l} {}_{i'=-k..k:\, i+i'\in \{1..n\},} \\ {}_{j'=-k..k:\, j+j'\in \{1..m\}} \end{array}} m_{i'j'} h_{i+i',j+j'}.
$$
For simplicity, we consider gray-scale images with a single channel.

The following examples illustrate how to construct filters of \our{} that produce identical results to classical convolution operations.
\begin{example} \label{ex1}
First, let us consider a linear convolution given by the mask
$$
\tensorsym{M}=\begin{bmatrix}
0 & 0 & 1 \\
0 & 0 & 0 \\
0 & 0 & 0
\end{bmatrix}.
$$
Observe that as the result of this convolution on the image $\tensorsym{H}$, every pixel is exchanged by its right upper neighbor, see Figure \ref{fig:ex1} (top). Now the image is represented as a graph, where the neighborhood $N_i$ of the pixel with coordinates $\vfor{i}=\begin{pmatrix}i_x\\i_y\end{pmatrix}$ is given by the pixels with coordinates $\vfor{j}=\begin{pmatrix}j_x\\j_y\end{pmatrix}$ such that $\vfor{j}-\vfor{i} \in \{-1,0,1\}^2$.

Given a vector\footnote{For simplicity, we consider an image with a single channel.} $\vfor{u} \in \R^2$ and a bias $b \in \R$ the (intermediate) graph operation is defined by
$$
g_i=\bar h_i(\vfor{u},b)=\sum_{j \in N_i}\relu{\left(\vfor{u}^T\left[\begin{pmatrix}j_x\\j_y\end{pmatrix} - \begin{pmatrix}i_x\\i_y\end{pmatrix}\right]+b\right)} \cdot h_j.
$$
Consider now the case when $\vfor{u}=2 \cdot \1, b=-3, \text{ where }\1=
(1,1)^T$. One can easily observe that
$$
\relu{(\vfor{u}^T \vfor{z} -b)}= \left\{
\begin{array}{ll}
0,& \text{ for } \vfor{z} \neq \1\\
1,& \text{ for } \vfor{z} = \1,
\end{array}
\right. 
$$
where $\vfor{z} = \vfor{j}-\vfor{i} \in \{-1,0,1\}^2$.

Consequently, we obtain that $g_i=h_{i+\1}$, which equals the result of the considered linear convolution.
\end{example}

\begin{example} \label{ex2}
Now, let us consider the mask, see Figure \ref{fig:ex1} (bottom):
$$
\tensorsym{M}=\begin{bmatrix}
0 & 0 & 1 \\
0 & 0 & 0 \\
1 & 0 & 0
\end{bmatrix}.
$$
This convolution cannot be obtained from graph representation using a single transformation as in previous example. 

To formulate this convolution, we define two intermediate operations for $k=1,2$:
$$
\bar{h}_i(\vfor{u}^T_k,b_k)= \sum_{j\in N_i} \relu{\left(\vfor{u}_k \left[\begin{pmatrix}j_x\\j_y\end{pmatrix} - \begin{pmatrix}i_x\\i_y\end{pmatrix}\right] + b_k\right) \cdot h_j}.
$$
where $\vfor{u}_1 = 2 \cdot \1, b_1= -3$ and $\vfor{u}_2 = -2 \cdot \1, b_2= -3$. 
The first operation extracts the right upper corner, while the second one extracts the left bottom corner, i.e.
$$
\bar{h}_i(\vfor{u}_1,b_1) = 1 \cdot h_{i+\1} \text{ and } \bar{h}_i(\vfor{u}_2,b_2) = 1 \cdot h_{i-\1}.
$$

Finally, we put 
$$
\bar{\tensorsym{H}}_i = \begin{bmatrix} \bar{h}_i(\vfor{u}_1,b_1)\\ \bar{h}_i(\vfor{u}_2,b_2)\end{bmatrix} = \begin{bmatrix} h_{i+\1}\\ h_{i-\1}\end{bmatrix}.
$$
Making an additional linear transformation defined by \eqref{eq:mlp} with $\vfor{w}_i = (1, 1)^T$, we obtain:
$$
g_{i}=\vfor{w}_i^T  \bar{\tensorsym{H}}_i = 1 \cdot h_{i+\1} + 1 \cdot  h_{i-\1}.
$$
\end{example}

Following the above examples, to obtain the result of applying an arbitrary $3 \times 3$ filter, we may need at most 9 operations using \eqref{eq:neigh-our} (one for replicating each of 9 entries). \our{} is also capable of imitating larger filters, see \cite{spurek2019geometric} for details.

\section{Experiments}

In this section, we evaluate our model on two machine learning tasks and compare it with related approaches. 

\subsection{Reconstruction}

First, as a proof of concept, we take into account MNIST database and consider the problem of restoring corrupted images, in which a part of data is hidden. To prepare this task, for each image of the size $28 \times 28$, we remove a square patch of the size $13 \times 13$. The location of the patch is uniformly sampled for each image. 

The reconstruction models are instantiated using the auto-encoder architecture (AE). In the case of our model, the encoder is implemented as \our{} with 5 spatial graph convolutional layers while the decoder is a typical deconvolutional neural network, which returns the image in the form of tensor. We emphasize that graph neural network is only needed for initial stage of processing (encoder) to avoid replacing missing values. Subsequent stages, e.g. decoder, can be implemented using typical non-graph networks. The result of our model is compared to CNNs combined with typical imputation techniques: (i) {\bf mask}, which is a zero imputation with an additional binary channel indicating unknown pixels (ii) {\bf mean} imputation, where absent attributes are replaced by mean values for a given coordinate (iii) {\bf k-nn} imputation, which substitutes missing features with mean values of those features computed from the k-nearest training samples (we use $k= 5$). For a fair comparison, every architecture (\our{} and CNNs) has the same structure, i.e. the number of layers and filters as well as the type of regularization.

\begin{figure}[htb!]
    \centering
    \includegraphics[width=0.9\textwidth]{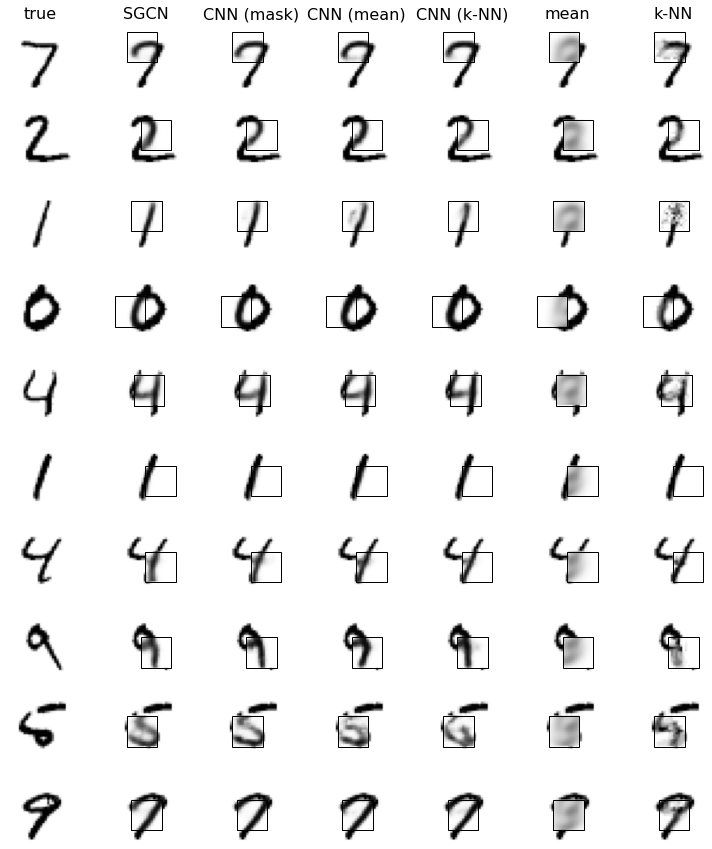}
    \caption{Reconstructions obtained for MNIST dataset (the first 10 images of test set). To demonstrate the influence of initial imputations on the final reconstructions returned by CNNs, we show the results of applying mean and k-NN imputations (last two columns). }
    \label{fig:my_label}
\end{figure}

We assume that the complete data are not available in the training phase. Therefore, for all models, the loss is defined as the mean-square error (MSE) calculated outside the missing region. This makes the problem more difficult than a typical inpainting task. To isolate the effect of processing incomplete images and directly compare \our{} with CNNs, we do not introduce additional losses and use only MSE in training.

\begin{table}[t]
    \centering
    \caption{Performance  on  three  machine learning tasks (lower is better). The first row shows mean-square error for reconstructing incomplete MNIST images, while the last two rows present test errors for classifying incomplete MNIST and SVHN images.}
    \label{tab:class}
    \begin{tabular}{ccccccc}
        \toprule
        Dataset & \our{} & GCN & CNN (mask) & CNN (mean) & CNN (k-NN) \\
        \midrule
        MNIST (MSE) & 0.0755 & - & 0.0760 & 0.0787 & {\bf 0.0725}\\
        MNIST (Error)& {\bf 4.6\%} & 31.4\% & 4.9\% & 5.9\% & 5.7\% \\ 
        SVHN (Error) & {\bf 16.6\%} & 74.6\% & 18.6\% & 19.9\% & 22.4\%\\
        \bottomrule
    \end{tabular}
\end{table}

It can be seen from the Figure \ref{fig:my_label} that \our{} gives similar results to CNN (mask). The reconstructions coincide on average with ground-truth and are free of artifacts. There was a problem in restoring digit "9" (last row), but the same holds for other methods. The results produced by CNN (mean) and CNN (k-NN) are sometimes blurry. To support the visual inspection with quantitative assessment, we calculate MSE inside the missing region, see the first row of Table~\ref{tab:class}. Surprisingly, CNN (k-NN) gives the highest resemblance with ground-truth in terms of MSE. While the reconstructions look visually less plausible than the ones returned by \our{} and CNN (mask), the pixel-wise agreement with ground-truth is higher. \our{} gives the second best result. It is important to observe the influence of initial imputation on the performance of CNNs. Since the k-NN imputation is significantly better than the mean imputation\footnote{k-NN imputation obtains MSE=0.0807 while mean imputation gives MSE=0.1265.}, the corresponding CNN model is able to restore input image more reliably. However, it is evident that mistakes made by k-NN imputation also negatively affect the performance of CNN (3rd and 9th rows). In contrast, our method is more stable, because it does not depend on imputation strategy. In consequence, it may give worse results than CNN when it is easy to predict missing values, but, at the same time, it should perform better if the imputation problem is more difficult. Another advantage is that \our{} is trained end-to-end (no preprocessing of missing values).


\subsection{Classification}

In the second experiment, we consider the classification task, in which incomplete data appear in both train and test phase. In addition to gray-scale handwritten digits retrieved from MNIST database, we also use color house-number images of the SVHN dataset. In the case of SVHN images of the size $32 \times 32$, we use patches of the size $15 \times 15$.

For a comparison, we use CNN models combined with the same imputation techniques as before, but, additionally, we consider ``vanilla'' {\bf GCN}~\cite{kipf2016semi}, which is one of the simplest GCNs that ignores spatial coordinates\footnote{We also experimented with graph attention network~\cite{velivckovic2017graph}, but the results did not improve.}. Classification network is composed of 8 convolutional layers. Each one contains 64 filters of the size $3 \times 3$. Batch normalization is used after every convolutional layer. As mentioned, we use analogical architecture for both graph convolutions and typical image convolutions.

It is evident from Table \ref{tab:class} (2nd and 3rd rows) that \our{} performs significantly better than the other version of GCN. It is not surprising because, in contrast to typical GCNs, \our{} introduces information about spatial coordinates to the model. Next observation is that \our{} gives lower errors than CNNs combined with imputation strategies. While the advantage of \our{} over the second best method in the case of MNIST is slight, the difference in accuracy is higher in the case of SVHN, which is significantly harder dataset to classify. As mentioned earlier, it may be difficult to reliably predict missing values for hard tasks, which, in consequence, negatively affects CNN models applied to completed data. Moreover, it can be seen that k-NN imputation is not so beneficial in classification problems as in reconstruction task -- CNN (k-NN) performs even worse than CNN (mean) on SVHN. As can be seen, the knowledge about missing pixels is more important for the success of classification CNNs than using specific imputation technique\footnote{We verified that combining masking with mean/k-nn imputation does not lead to further improvement of CNNs.}. In contrast to CNN (mask), which uses an additional binary channel to pass the information about unknown values to the neural model, \our{} directly ignores missing pixels, which is more natural.

\section{Conclusion}

We presented an alternative way of learning neural networks from incomplete images, which does not require replacing missing values at the preprocessing stage. While graph representation of incomplete images avoids the problem of imputation, applying \our{} allows us to reflect the action of classical CNNs. Our model is trained end-to-end without any missing data preprocessing (imputation) as in the case of CNNs. Since the proposed model completely ignores the information about missing values, it is especially useful in the case of complex tasks, where imputation strategies introduce noise and unreliable information.
The main disadvantage of our approach is the computational cost of using GCNs. In contrast to classical CNNs, the current implementations of GCNs are less efficient and it is difficult to manage large graphs created from high dimensional images.

\section*{Acknowledgements}

The work of M. \'Smieja was supported by the National Science Centre (Poland) grant no. 2017/25/B/ST6/01271. 
The work of \L{}. Struski was supported by the Foundation for Polish Science Grant No. POIR.04.04.00-00-14DE/18-00 co-financed by the European Union under the European Regional Development Fund. 
The work of P. Spurek  was supported by the National Centre of Science (Poland) Grant No. 2019/33/B/ST6/00894. 
The work of \L{}. Maziarka was supported by the National Science Centre (Poland) grant no. 2018/31/B/ST6/00993. 

\bibliographystyle{splncs04}
\bibliography{main}

\begin{thebibliography}{10}
\providecommand{\url}[1]{\texttt{#1}}
\providecommand{\urlprefix}{URL }
\providecommand{\doi}[1]{https://doi.org/#1}

\bibitem{chechik2008max}
Chechik, G., Heitz, G., Elidan, G., Abbeel, P., Koller, D.: Max-margin
  classification of data with absent features. Journal of Machine Learning
  Research  \textbf{9},  1--21 (2008)

\bibitem{danel2020processing}
Danel, T., {\'S}mieja, M., Struski, {\L}., Spurek, P., Maziarka, L.: Processing
  of incomplete images by (graph) convolutional neural networks. In: ICML
  Workshop on The Art of Learning with Missing Values (Artemiss). p.~6 (2020)

\bibitem{spurek2019geometric}
Danel, T., Spurek, P., Tabor, J., {\'S}mieja, M., Struski, {\L}., S{\l}owik,
  A., Maziarka, {\L}.: Spatial graph convolutional networks. arXiv preprint
  arXiv:1909.05310  (2020)

\bibitem{dekel2010learning}
Dekel, O., Shamir, O., Xiao, L.: Learning to classify with missing and
  corrupted features. Machine Learning  \textbf{81}(2),  149--178 (2010)

\bibitem{globerson2006nightmare}
Globerson, A., Roweis, S.: Nightmare at test time: robust learning by feature
  deletion. In: Proceedings of the International Conference on Machine
  Learning. pp. 353--360. ACM (2006)

\bibitem{goldberg2010transduction}
Goldberg, A., Recht, B., Xu, J., Nowak, R., Zhu, X.: Transduction with matrix
  completion: Three birds with one stone. In: Advances in neural information
  processing systems. pp. 757--765 (2010)

\bibitem{goodfellow2016deep}
Goodfellow, I., Bengio, Y., Courville, A.: Deep learning. MIT press (2016)

\bibitem{grangier2010feature}
Grangier, D., Melvin, I.: Feature set embedding for incomplete data. In:
  Advances in Neural Information Processing Systems. pp. 793--801 (2010)

\bibitem{hazan2015classification}
Hazan, E., Livni, R., Mansour, Y.: Classification with low rank and missing
  data. In: Proceedings of The 32nd International Conference on Machine
  Learning. pp. 257--266 (2015)

\bibitem{kipf2016semi}
Kipf, T.N., Welling, M.: Semi-supervised classification with graph
  convolutional networks. arXiv preprint arXiv:1609.02907  (2016)

\bibitem{lecun1998gradient}
LeCun, Y., Bottou, L., Bengio, Y., Haffner, P.: Gradient-based learning applied
  to document recognition. Proceedings of the IEEE  \textbf{86},  2278--2324
  (1998)

\bibitem{liu2018image}
Liu, G., Reda, F.A., Shih, K.J., Wang, T.C., Tao, A., Catanzaro, B.: Image
  inpainting for irregular holes using partial convolutions. In: Proceedings of
  the European Conference on Computer Vision (ECCV). pp. 85--100 (2018)

\bibitem{mcknight2007missing}
McKnight, P.E., McKnight, K.M., Sidani, S., Figueredo, A.J.: Missing data: A
  gentle introduction. Guilford Press (2007)

\bibitem{netzer2011reading}
Netzer, Y., Wang, T., Coates, A., Bissacco, A., Wu, B., Ng, A.Y.: Reading
  digits in natural images with unsupervised feature learning. In: NIPS
  Workshop on Deep Learning and Unsupervised Feature Learning (2011)

\bibitem{pelckmans2005handling}
Pelckmans, K., De~Brabanter, J., Suykens, J.A., De~Moor, B.: Handling missing
  values in support vector machine classifiers. Neural Networks
  \textbf{18}(5),  684--692 (2005)

\bibitem{smieja2019generalized}
{\'S}mieja, M., Struski, {\L}., Tabor, J., Marzec, M.: Generalized rbf kernel
  for incomplete data. Knowledge-Based Systems  \textbf{173},  150--162 (2019)

\bibitem{smieja2018processing}
{\'S}mieja, M., Struski, {\L}., Tabor, J., Zieli{\'n}ski, B., Spurek, P.:
  Processing of missing data by neural networks. In: Advances in Neural
  Information Processing Systems. pp. 2719--2729 (2018)

\bibitem{velivckovic2017graph}
Veli{\v{c}}kovi{\'c}, P., Cucurull, G., Casanova, A., Romero, A., Lio, P.,
  Bengio, Y.: Graph attention networks. arXiv preprint arXiv:1710.10903  (2017)

\bibitem{xia2017adjusted}
Xia, J., Zhang, S., Cai, G., Li, L., Pan, Q., Yan, J., Ning, G.: Adjusted
  weight voting algorithm for random forests in handling missing values.
  Pattern Recognition  \textbf{69},  52--60 (2017)

\end{thebibliography}

\end{document}